\documentclass[10pt,twocolumn,letterpaper]{article}
\usepackage{tabularx}
\usepackage[utf8]{inputenc}
\usepackage[table,xcdraw,dvipsnames]{xcolor}
\usepackage[normalem]{ulem}
\useunder{\uline}{\ul}{}
\usepackage{iccv}
\usepackage{times}
\usepackage{epsfig}
\usepackage{graphicx}
\usepackage{amsmath}
\usepackage{amssymb}
\usepackage{bm}
\usepackage{booktabs}
\usepackage{float}
\usepackage{graphicx}
\usepackage{algorithmic}
\usepackage{algorithm}
\usepackage{amsmath}
\usepackage{textcomp}
\usepackage{amsthm}
\usepackage{amsmath}
\usepackage{amssymb}
\usepackage{times}
\usepackage{epsfig}
\usepackage{amssymb}
\usepackage{multirow}
\usepackage{url,graphicx,color,xcolor,booktabs,colortbl,threeparttable}
\usepackage{pifont}
\usepackage{enumitem}
\usepackage{array}
\usepackage{bm}
\usepackage{url}
\usepackage{ifpdf}
\usepackage{ifxetex}

\usepackage[breaklinks=true,bookmarks=false]{hyperref}

\iccvfinalcopy 

\def\iccvPaperID{****} 

\ificcvfinal\fi

\begin{document}
\title{ChartReader: A Unified Framework for Chart Derendering and Comprehension without Heuristic Rules}
\vspace{-2mm}
\author{Zhi-Qi Cheng\textsuperscript{$1$}\qquad
Qi Dai\textsuperscript{$2$}\qquad
Siyao Li\textsuperscript{$1,3$}\qquad
Jingdong Sun\textsuperscript{$1$} \qquad\\
Teruko Mitamura\textsuperscript{$1$}\qquad
Alexander G. Hauptmann\textsuperscript{$1$}\\
\textsuperscript{1}{Carnegie Mellon University}\qquad
\textsuperscript{2}{Microsoft Research}\qquad
\textsuperscript{3}{Amazon}\qquad\\
{\tt \small \{zhiqic,alex,teruko\}@cs.cmu.edu, {qid@microsoft.com},
\{siyaol,jingdons\}@andrew.cmu.edu}}
\vspace{-2mm}
\maketitle
\ificcvfinal\thispagestyle{empty}\fi

\begin{abstract}
\vspace{-1mm}
Charts are a powerful tool for visually conveying complex data, but their comprehension poses a challenge due to the diverse chart types and intricate components. Existing chart comprehension methods suffer from either heuristic rules or an over-reliance on OCR systems, resulting in suboptimal performance.~To address these issues, we present ChartReader, a unified framework that seamlessly integrates chart derendering and comprehension tasks. Our approach includes a transformer-based chart component detection module and an extended pre-trained vision-language model for chart-to-X tasks.~By learning the rules of charts automatically from annotated datasets, our approach eliminates the need for manual rule-making, reducing effort and enhancing accuracy.~We also introduce a data variable replacement technique and extend the input and position embeddings of the pre-trained model for cross-task training. We evaluate ChartReader on Chart-to-Table, ChartQA, and Chart-to-Text tasks, demonstrating its superiority over existing methods. Our proposed framework can significantly reduce the manual effort involved in chart analysis, providing a step towards a universal chart understanding model.~Moreover, our approach offers opportunities for plug-and-play integration with mainstream LLMs such as T5 and TaPas, extending their capability to chart comprehension tasks.\footnote{ Code is at \url{https://github.com/zhiqic/ChartReader}}
\end{abstract}

\vspace{-4mm}
\section{Introduction}
\label{sec:introduction}
\vspace{-1mm}
\begin{figure} [!ht]
\small
\begin{center}
    \includegraphics[width=0.8\linewidth]{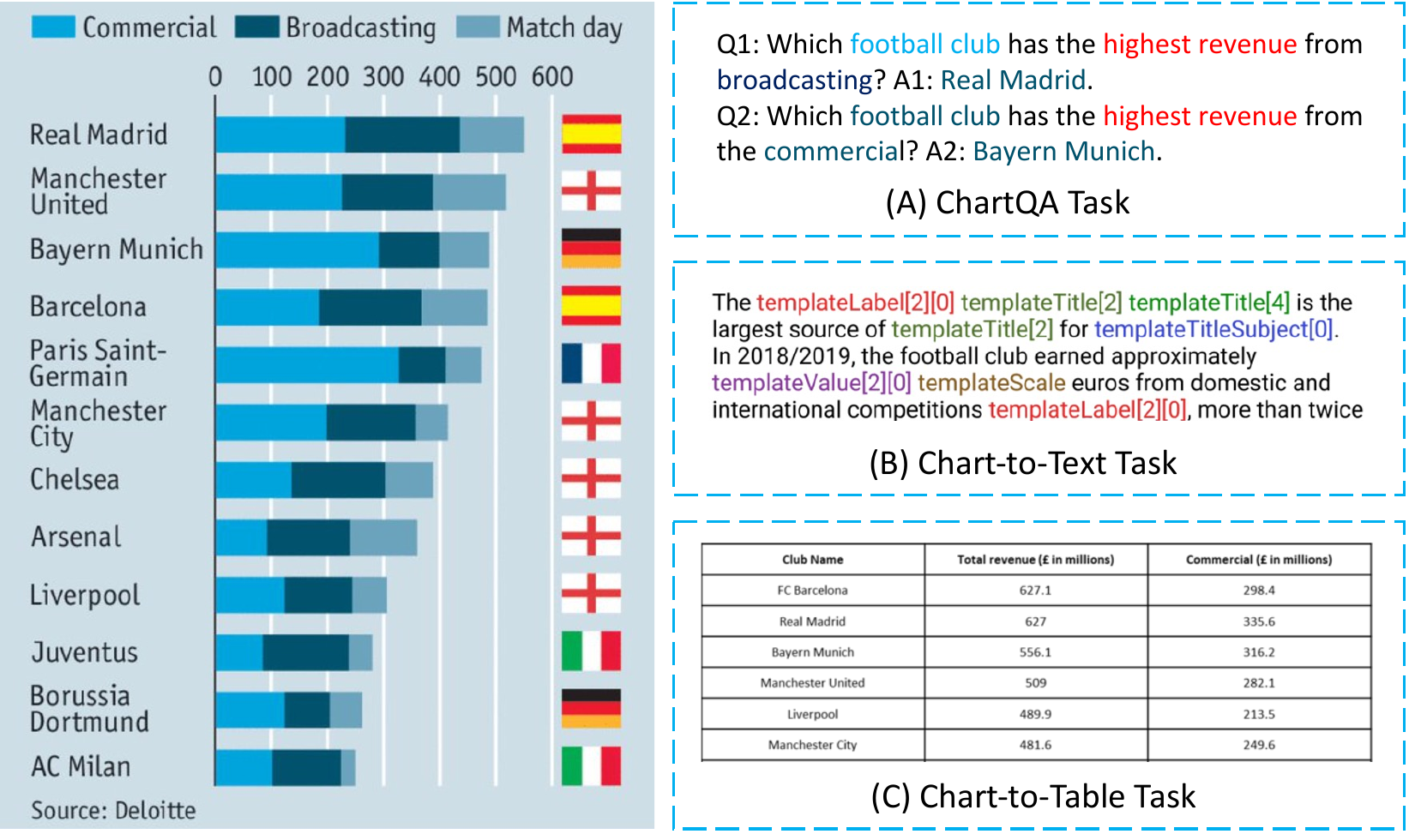}
\end{center}
\vspace{-3mm}
\caption{\small Illustration of chart derendering and comprehension tasks.~The Chart-to-Table task aims to transform a chart into a machine-readable table, while ChartQA and Chart-to-Text tasks involve answering questions and summarizing the content of the chart, respectively.~[Best viewed in color].}
\label{fig:1}
 \vspace{-6mm}
\end{figure}	

The adage, ``a picture is worth a thousand words," underscores the immense value of charts found on various websites and articles, which often depict knowledge that cannot be conveyed through words alone.~Chart derendering, which refers to the conversion of charts into tables (i.e., Chart-to-Table~\cite{choi2019visualizing, liu2019data, luo2021chartocr}), is widely viewed as essential in facilitating a range of downstream tasks, such as chart question-answering (ChartQA~\cite{kafle2020answering,methani2020plotqa,zou2020affinity}) and chart summarization (Chart-to-Text~\cite{chen2019figure, demir2012summarizing, obeid2020chart}). As shown in Figure~\ref{fig:1}, the Chart-to-Table task aims to recognize the chart as a machine-readable table, while ChartQA and Chart-to-Text tasks involve answering pre-set questions and summarizing the content of the chart, respectively. The interdependence and mutual value of these research tasks have also been emphasized in earlier studies~\cite{ huang2021visual,Wang2022, wang2021survey}, underscoring their critical role in chart comprehension research.

\begin{figure*}[t]
\small
\centering
\includegraphics[width=0.85\linewidth]{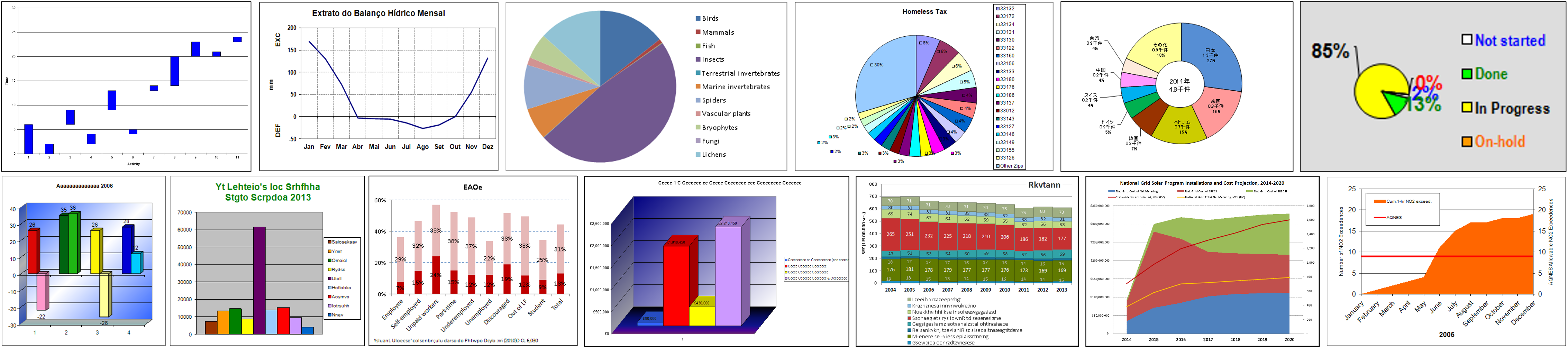}
\caption{\small Demonstrates the complexity of charts in EC400K~\cite{luo2021chartocr}, which can vary in type, design, and visual properties. Charts can contain intricate components, texture variations, and speckled backgrounds, posing challenges for chart comprehension.~[Best viewed in color].}
\label{fig:2}
\vspace{-4mm}
\end{figure*}

Despite the critical role of chart comprehension, existing research has failed to address the three sub-tasks separately, let alone propose a universal solution. As depicted in Figure~\ref{fig:2}, charts come in various types, each designed to convey domain-specific knowledge, and can exhibit intricate components, texture variations, and speckled backgrounds. Confronted with such complex charts, existing chart comprehension methods face two main problems.

Firstly, existing chart derendering approaches~\cite{balaji2018chart, choi2019visualizing,gao2012view, liu2019data, luo2021chartocr}~(i.e.,~Chart-to-Table) resort to heuristic rules that demand extensive domain knowledge and effort to formulate.~For example,~ChartOCR~\cite{luo2021chartocr}, a pioneering method, requires chart classification to identify categories first and then detects different components using various pre-defined heuristic rules.~To avoid complicated rule-making, some methods even try only one set of limited chart types, such as bar~\cite{Daggubati2022, Rane2021} and line~\cite{Kato2022,Mahinpei2022} charts. These limitations impede the ability to extract data for unknown categories. As a result, many latest methods~\cite{liu2022matcha,methani2020plotqa} even use tables directly from ground truth to complete the answers and summary tasks. It is evident that these studies are challenging to automate and struggle to extract data from real-world charts.
	
On the other hand, current chart comprehension methods, such as Chart-to-Text~~\cite{chen2019figure, demir2012summarizing, obeid2020chart} and ChartQA~\cite{Kantharaj2022,levy2022classification,masry2022chartqa,methani2020plotqa,singh2020stl}, often heavily rely on off-the-shelf OCR systems or pre-extracted tables from the ground truth. By treating chart derendering as a black box, these methods neglect the visual and structural information of the charts, resulting in the following issues: ~(1)~Chart-to-Text and ChartQA tasks devolve into text-only quizzes~\cite{liu2022deplot,masry2022chartqa}, as they cannot extract visual semantics from chart derendering.~This explains why OCR-based and end-to-end methods, such as LayoutLM~\cite{xu2020layoutlm}, PresSTU~\cite{kil2022prestu}, PaLI~\cite{chen2022pali}, CoCa~\cite{yu2022coca}, Donut~\cite{kim2022ocr}, and Dessurt~\cite{davis2023end}, have shown suboptimal results in chart understanding.~(2)~Chart-to-table tasks do not benefit from chart comprehension tasks.~Due to the lack of understanding to the visual semantics in charts, existing systems, such as those using OCR systems~\cite{liu2022deplot,luo2021chartocr}, struggle to accurately convert charts to tables.~Overall, we contend that the problems with chart comprehension arise from an over-reliance on predefined rules and a lack of a universal framework to support multi-tasking.

In light of the previous analysis, it seems that a visual-language model is a promising direction for building a universal framework.~However, while Pix2Struct~\cite{lee2022pix2struct}, a pretraining strategy for visually-situated language, has shown superior performance over OCR-based models~\cite{appalaraju2021docformer,biten2022latr,li2021selfdoc,wang2022git}, it is not suitable for chart derendering. 
Moreover, despite recognizing this issue, Metcha~\cite{liu2022matcha} had to rely on Pix2Struct as a backbone due to the lack of better visual-language models. Nonetheless, neither Pix2Struct nor Metcha address the two issues identified earlier:~1) excessive reliance on heuristic rules in table derendering, and 2) heavy reliance on existing OCR systems despite attempts to incorporate additional chart comprehension tasks.

To overcome these concerns, we introduce ChartReader, a unified framework that seamlessly integrates chart derendering and comprehension tasks. Our approach comprises a rule-free chart component detection module and an extended pre-trained vision-language model for chart-to-X (text/table/QA) tasks. Unlike heuristic rule-based methods, our approach leverages a transformer-based approach to detect the location and type of chart components, enabling automatic rule learning by processing existing annotated datasets. To enhance cross-task training, we extend the input and position embeddings of the pre-trained model and introduce a data variable replacement technique.
Specifically, we standardize chart-to-X (table/text) tasks as question-answering problems, allowing us to solve multiple chart understanding tasks effectively. 
Additionally, the model generates the data variable instead of the actual value to avoid errors and hallucinations, which improves the consistency in multi-task training.
Our approach represents a step towards a unified chart understanding model, as validated through experiments. The proposed framework has the potential to reduce the manual effort involved in chart analysis, paving the way for more efficient and accurate chart comprehension.

To summarize, our contributions are:~1)~a unified framework that seamlessly integrates chart derendering and comprehension tasks;~2)~a rule-free chart component detection module that leverages a transformer-based approach to automatically learn the rules of charts;
3)~extending the input and position embeddings of the pre-trained LLMs and employing a data variable replacement technique to improve cross-task training; 4)~validating our approach through experiments, demonstrating significant improvement over existing methods in chart understanding tasks. 

\vspace{-1mm}
\section{Related Works}
\vspace{-1mm}
This section reviews related works on chart understanding, including chart derendering, question-answering, and summarization, highlighting the differences from previous work and the impact on visual-language research. 

\noindent \textbf{Chart Derendering}, also known as Chart-to-Table, involves identifying the constituent components of an image of a chart, such as bars, pies, and legends, to extract the underlying data represented by the chart. Traditional methods~\cite{balaji2018chart, gao2012view,jung2017chartsense, poco2017reverse, savva2011revision} relied on hand-designed rules based on edges and colors, which were time-consuming and not easily generalized to new chart types. Deep-learning based approaches~\cite{choi2019visualizing, liu2019data, luo2021chartocr} utilizing object detection and text recognition networks have shown to detect chart components accurately with better generalizability. However, some recent work~\cite{liu2022deplot} has attempted to use pre-trained large-scale language models (LLMs) for plot-to-table task, without recognizing the structure and components of the chart. Despite the progress made, the flagship method, ChartOCR~\cite{luo2021chartocr}, still relies heavily on human rules and different networks for each chart type. In contrast, our approach eliminates the need for heuristic rules and can easily be extended to unseen chart styles.

\noindent \textbf{Chart Question-Answering}, or ChartQA, is the task of answering questions related to charts by utilizing both visual and textual information. Early methods~\cite{kahou2017figureqa} relied on relation networks to represent relationships between image objects or between visuals and questions, while dynamic encoding was introduced~\cite{kafle2018dvqa} to handle out-of-vocabulary candidate answers. Later approaches~\cite{kafle2020answering, zou2020affinity} have leveraged multi-modal models that use LSTM~\cite{hochreiter1997long} and DenseNet~\cite{huang2017densely} to extract visual and text features and fuse embeddings to answer questions.
Recent methods, such as OpenCQA~\cite{Kantharaj2022}, CRCT~\cite{levy2022classification}, ChartQA~\cite{masry2022chartqa}, PlotQA-M~\cite{methani2020plotqa}, and STL-CQA~\cite{singh2020stl}, have incorporated transformers to capture complex visual and text information when answering questions.
Furthermore, some methods~\cite{liu2022matcha,masry2022chartqa} have attempted to use pre-trained LLMs for chart and language data modeling, but they often disregard chart characteristics and rely solely on ground-truth tables. In contrast, our work unifies chart summarization and derendering tasks into the ChartQA task, using a single sequence-to-sequence framework that elegantly combines chart characteristics and meaningful aspects of different tasks into LLMs.

\noindent \textbf{Chart Summarization}, refers to Chart-to-Text, aims to generate natural language summaries from visual charts. Traditional approaches~\cite{ferres2013evaluating, mittal1998describing, reiter2007architecture} employed templates to provide brief descriptions of the chart's appearance, while others~\cite{fasciano2000intentions, green2004autobrief} used heuristics or planning-based methods to create multimedia presentations. Recently, Natural Language Generation (NLG) techniques based on heuristic rules have been used, including statistical analysis~\cite{cui2019datasite,srinivasan2018augmenting,wang2019datashot} to infer insights and encoder-decoder architectures~\cite{chen2019figure,demir2012summarizing,  obeid2020chart} to generate template-based captions. However, these methods have a common limitation in that they rely on predefined template-based approaches, which may lack generality and variation in grammatical style and lexical choices. In contrast, our research proposes a universal sequence-to-sequence chart understanding framework that utilizes data-driven models to generate more diverse and informative summaries.

\noindent \textbf{Advancements in Vision-Language Research}.~Recently, vision-language research has primarily focused on natural images, relying on visually-grounded reasoning~\cite{cheng2022gsrformer,liu2021visually,suhr2019nlvr2} and synthesized datasets~\cite{andreas2016neural,cheng2018learning,johnson2017clevr,suhr2017corpus} for evaluation. However, these works do not capture the complexities of real-world visual language, especially in chart understanding tasks. While OCR-based and end-to-end methods, such as LayoutLM~\cite{xu2020layoutlm}, PresSTU~\cite{kil2022prestu}, PaLI~\cite{chen2022pali}, 
CoCa~\cite{yu2022coca},
Donut~\cite{kim2022ocr}, Dessurt~\cite{davis2023end}, and Pix2Struct~\cite{lee2022pix2struct} have been developed for visually-situated language, they do not specifically address the challenges posed by chart understanding. Despite the widespread use of the encoder-decoder framework, it still requires specific design considerations, such as determining which inputs are valuable and eliminating artificial rules. Without these modules, existing models, such as LaTr~\cite{biten2022latr}, GIT~\cite{wang2022git}, DocFormer~\cite{appalaraju2021docformer}, and SelfDoc~\cite{li2021selfdoc}, cannot be directly applied to chart understanding. In contrast, our work eliminates these limitations by not requiring predefined heuristic rules and unifying data extraction and understanding in an encoder-decoder framework, achieving impressive results across multiple datasets.

\begin{figure*}
\small
\centering
\includegraphics[width=0.85\linewidth]{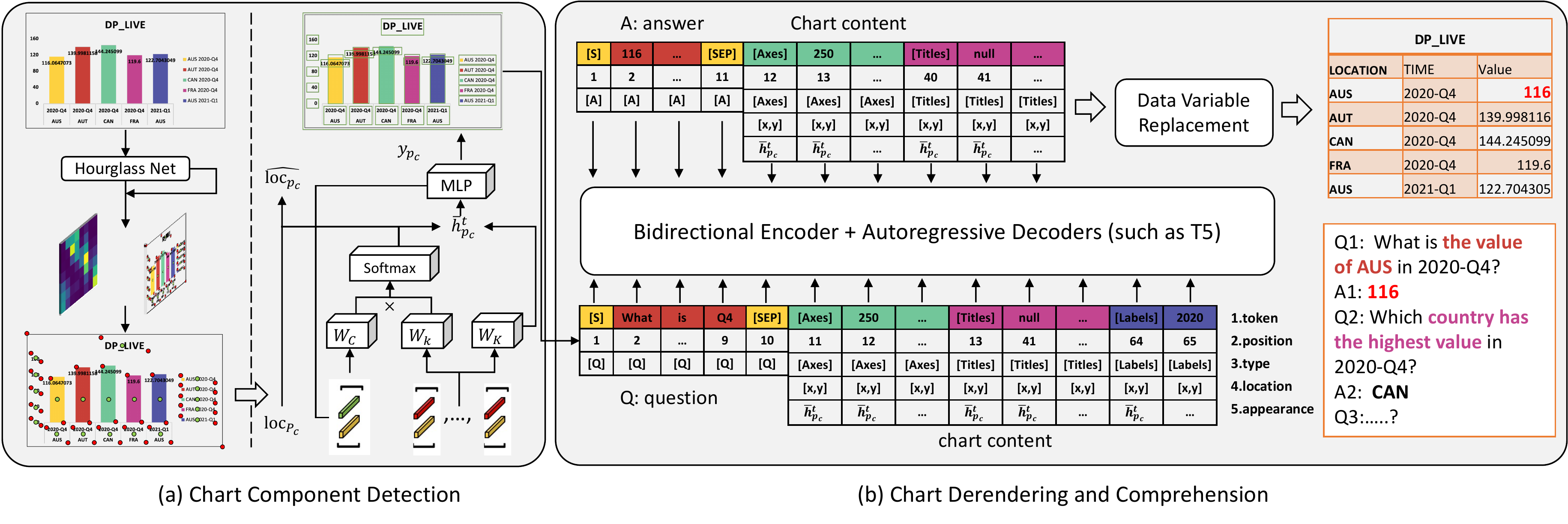}
\vspace{-1mm}
\caption{\small illustrates the components of our proposed unified framework for various chart analysis tasks, including chart-to-table, chart-to-text, and chartQA. The framework is comprised of two main modules: (1) a transformer-based chart component detection module that eliminates the need for manual rule-making, and (2) an extended pre-trained vision-language model that enables chart-to-X (text/table/QA) tasks. The combination of these two modules enables seamless integration and efficient execution of chart comprehension tasks.}
\label{fig:3}
\vspace{-3mm}
\end{figure*}

\vspace{-1mm}
\section{Unified ChartReader Framework}
\vspace{-1mm}
Our unified framework aims to support various chart analysis tasks, including chart-to-table, chart-to-text, and chartQA. As shown in Figure~\ref{fig:3}, it consists of two main components:~(1)~a rule-free chart component detection module, and~(2)~an extended pre-trained vision-language model for chart-to-X~(text/table/QA) tasks.

\vspace{-1mm}
\subsection{Chart Component Detection}
\vspace{-1mm}
We exploit a transformer-based approach to detect the location and type of chart components without relying on heuristic rules.~Our approach consists of three main steps:~1)~center/keypoint detection,~2)~center/keypoint grouping, and~3)~component position/type prediction.

\begin{figure} [!t]
\small
\begin{center}
    \includegraphics[width=0.8\linewidth]{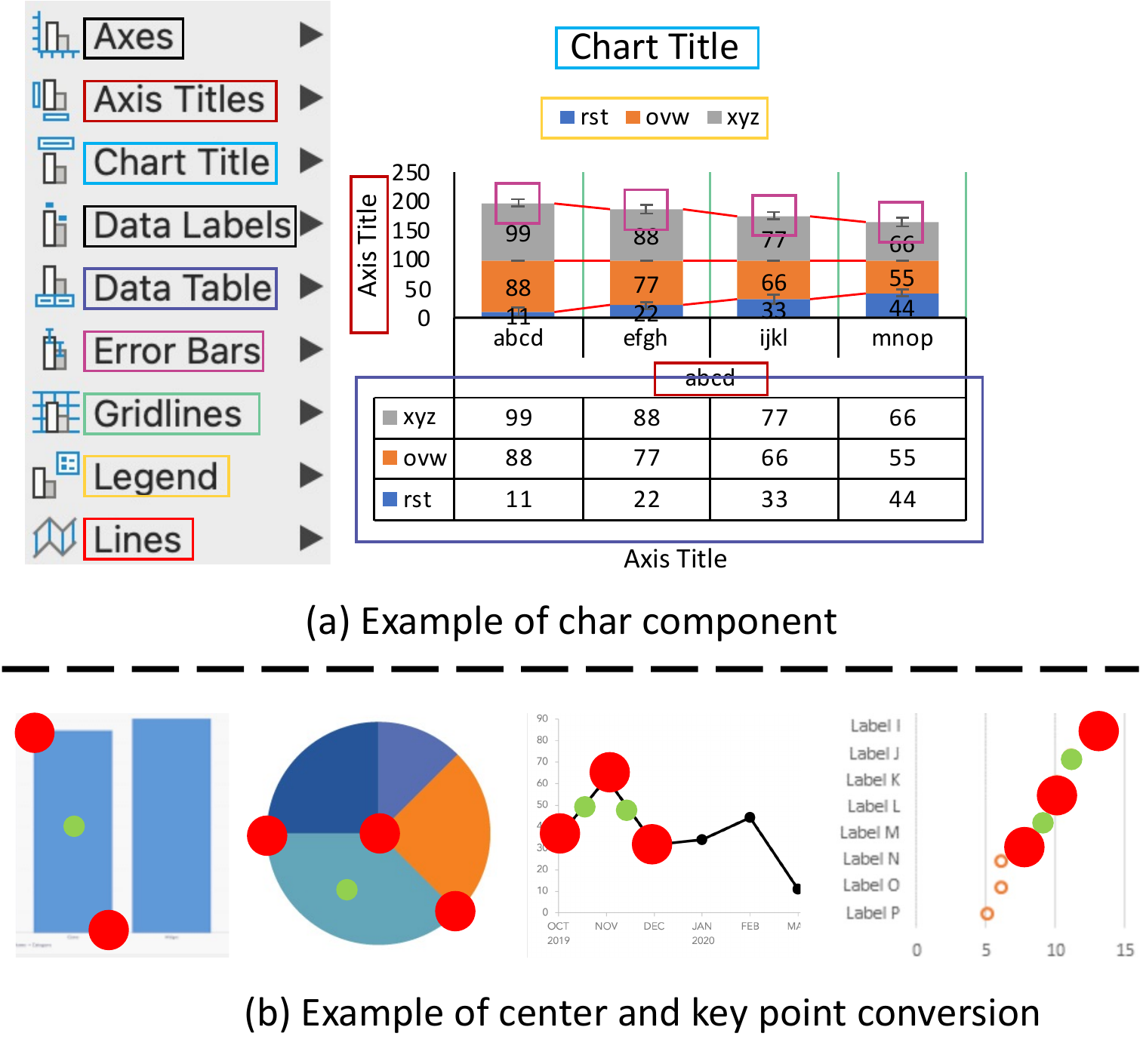}
\end{center}
\vspace{-4mm}
\caption{\small demonstrates the conversion of chart components into center and key points, which are inferred from existing datasets and used to locate chart components in our updated model. This approach eliminates the need for heuristic rules and allows the model to overcome the effects of style variations. Specifically, we convert the upper left and lower right corners of each component into center and key points, respectively.~[Best viewed in color].}
\label{fig:4}
 \vspace{-4mm}
\end{figure}

\noindent \textbf{Overcoming Heuristics}.~The motivation behind this is to overcome the limitations of heuristic rule-based methods in handling various chart styles that heavily rely on domain-specific knowledge and complex rule definitions. By processing annotated existing datasets, our model can automatically learn the rules of charts. Furthermore, the optimized framework can seamlessly be applied to other downstream chart understanding tasks.~As illustrated in Figure~\ref{fig:4}, our updated model relies on center and key points inferred from existing datasets instead of heuristic rules to locate chart components and overcome the effects of style variations. We convert the upper left and lower right corners and position centers of each component into the center and key points, respectively.~Although our approach still requires a large amount of annotated data, we do not need to design specific rules for each chart type. Our approach represents a step towards a unified chart understanding model and has been validated through experiments.

\noindent \textbf{Step-1:~Center/Keypoint Detection}.~We detect the centers $p_c \in \mathcal{C}$ and keypoints $p_k \in \mathcal{K}$ of chart components using the Hourglass network~\cite{newell2016stacked} without the corner pooling layer to improve generalization across chart styles. The location $\widehat{\text{loc}}_{p}\in \mathbb{R}^2$ and component type $\hat{y}_{p}\in \{1,...,\mathcal{T}\}$ for all centers and keypoints are predicted using the focal loss $L_{focal}$ \cite{law2018cornernet} and the smooth $L_1$ loss \cite{girshick2015fast}.~We use the central pooling layer for all component types, rather than different pooling strategies for each component type \cite{luo2021chartocr}, to effectively detect center and keypoint of chart components in various styles.

\noindent \textbf{Step-2:~Center/Keypoint Grouping}.~For each detected center and keypoint, we extract the features of their corresponding positions to obtain initial embeddings. To better determine the chart component, we introduce a type token $\phi_{p_{c}}$ and $\phi_{p_{k}}$ and use multi-head attention to obtain the weights of the center point $p_c$ and keypoint $p_k$ as,
\begin{equation}
\label{eq:1}
\centering
\mathcal{G}(p_c,p_k)=\left( \mathbf{W}_\mathcal{C} \left[h_{p_c},\phi_{p_c}\right] \right)^{T}
\mathbf{W}_\mathcal{K}
\left[h_{p_k},\phi_{p_k}\right],
\vspace{-1mm}
\end{equation}
where $h_{p_c}$ and $h_{p_k}$ respectively represent the hidden features of the center point and keypoint. The matrices $\mathbf{W}_\mathcal{C}$ and $\mathbf{W}_\mathcal{K}$ are projection matrices of the hidden features of the center point and the key point. We use fixed sinusoidal features to encode the absolute x- and y-axis positions and add $\phi_{p_{c}}$ and $\phi_{p_{k}}$ to the embedding before multiplication. After obtaining the weights of $\mathcal{G}$ using Eqn.~\ref{eq:1}, we compute the final grouping score by normalizing the weights with a softmax function over the entire set of keypoints as,
\begin{equation}
\label{eq:2}
\centering
\text{attn}(p_c, p_k) =\frac{\exp{(\mathcal{G}(p_c,p_k))}}{\sum_{\bar{k}\in \mathcal{K}} \exp ({\mathcal{G}(p_c, p_{\bar{k}})})},
\vspace{-1mm}
\end{equation}
where the softmax function sorts keypoints $\bar{k} \in \mathcal{K}$ from the same component and filters the most relevant ones (i.e.,~$p_{k}$) for each center point $p_c$.~This approach enables effective grouping of centers and keypoints to their corresponding chart components.

\noindent \textbf{Step-3:~Component Position \& Type Prediction}.~We predict the position and type of each chart component by optimizing the center point position using the grouping score and keypoint positions from the Hourglass network.~We compute the weighted average of the keypoint positions $\widehat{\text{loc}}_{p_k}$ for each center point $p_c$ using the grouping score as,
\begin{equation}
\label{eq:3}
\widehat{\text{loc}}_{p_c} = \sum_{k \in \mathcal{K}}
\text{attn}(p_c, p_k)\widehat{\text{loc}}_{p_k}.
\vspace{-1mm}
\end{equation}
To ensure that the predicted positions are close to the ground truth positions $\text{loc}_{p_c}$, we use the location loss $L_{loc}$, which is defined as,
\begin{equation}
\label{eq:4}
\centering
L_{loc}=\sum_{c\in \mathcal{C}}
|\widehat{\text{loc}}_{p_c}- \text{loc}_{p_c}|.
\vspace{-1mm}
\end{equation}
This supervises the optimization process and helps to accurately predict the position of each chart component.

To predict the type of chart component, we compute a weighted sum of the keypoint embeddings using the grouping score as,
\begin{equation}
\label{eq:5}
\bar{h}_{p_c}=\sum_{k \in \mathcal{K}} \text{attn}(p_c, p_k)\mathbf{W}_{\mathcal{K}}[h_{p_k}, \phi_{p_k}].
\vspace{-1mm}
\end{equation}
The resulting center point embedding $\bar{h}_{p_c}$ is then passed through an MLP layer to obtain the predicted probability distribution over all component types, which is then compared with the ground truth component type labels $y_{p_c}$ using the cross-entropy loss $L_{\text{CLS}}$ as,
\begin{equation}
\label{eq:6}
\centering
L_{\text{CLS}}=-\sum_{c \in \mathcal{C}}
y_{p_c} \log (\text{softmax}(\bar{h}_{p_c})).
\vspace{-1mm}
\end{equation}
This supervises the optimization process and helps to accurately predict the type of each chart component.

\vspace{-1mm}
\subsection{Chart Derendering and Comprehension}
\vspace{-1mm}
We propose a unified chart understanding framework that handles Chart-to-X (text/table/QA) tasks as chartQA tasks, as illustrated in Figure X. To improve cross-task training, we extend~(1)~the input and position embeddings of the pre-trained model and~(2)~employ a data variable replacement technique.

\noindent \textbf{Motivation \& Reasoning}.~Our approach is motivated by two main reasons:
Firstly, we treat both the chart-to-X (table/text) tasks as question-answering problems.~Specifically, for chart-to-table, the combination of axis labels and legends forms the question, and the extracted components serve as the answer. Similarly, chart-to-text is treated as a Q\&A task to fill in blank templates generated by removing redundant information. By standardizing these previously independent tasks as Q\&A tasks, we can effectively train and solve multiple chart understanding tasks.
Secondly, recent advancements in these tasks have adopted sequence-to-sequence models.~To address all of them in one framework, we extend pre-trained LLMs to diverse chart comprehension tasks. However, integrating these tasks is challenging, and thus, we conduct numerous experiments to determine how to extend input and position encoding. Additionally, we propose a data variable replacement technique to enhance the consistency of multi-task training. Our findings provide possibilities for extending pre-trained LLMs to chart comprehension, and it can be seamlessly extended to mainstream LLMs such as T5 and TaPas.

\noindent \textbf{Positional \& Input Embedding}.~To support tasks such as Chart-to-Table, Chart-to-Text, and ChartQA, we have extended the input and positional embeddings from natural language to chart-style data.~The fused embedding $z_k$ at the $k$-th position of the sequence is defined as,
\begin{equation}
z_k = {\rm LN}(z^{{\rm pos}}_k+(z^{{\rm token}}_k + z^{{\rm typ}}_k + z^{{\rm loc}}_k + z^{{\rm app}}_k)),
\end{equation}
where ${\rm LN}()$ denotes layer normalization~\cite{ba2016layer}.~Firstly, the positional embedding $z^{\rm pos}_k$ is set to zero, as modern vision-language pre-trained models use relative position embeddings.~Secondly, the input embedding includes the type $z^{\rm typ}_k$, location $z^{\rm loc}_k$, and appearance $z^{\rm app}_k$ of chart components, as well as the token $z^{\rm token}_k$ obtained from other textual information in the dataset.

Specifically, the $z^{\rm token}_k$ is generated by tokenizing the concatenation of textual words and is denoted as,
\begin{equation}
\nonumber
\footnotesize
\centering
x^{\rm token}= \biggl\{
\begin{aligned}
&{\rm \texttt{[S]}}, {\rm [w_1]}, ..., {\rm [w_m]}, {\rm \texttt{[SEP]}}, [{\rm y_{p_c}^1}], w_{1,1}\\&,..., w_{1,m}, 
[{\rm y_{p_c}^2}], \dots,
[{\rm y_{p_c}^N}], 
w_{r_N,1}, ..., w_{r_N,m},...,.
\end{aligned}
\biggr\},
\end{equation}
where~${\rm\texttt{[S]}}$ is used to distinguish between questions and answers, and ${\rm \texttt{[SEP]}}$ indicates the presence of chart context. To incorporate information about chart components, a special token $[\rm {y_{p_c}^i}]$ is introduced to represent the chart component type $y_{p_c}^i \in \{1,...,\mathcal{T}\}$ (such as $[$Axes$]$; see Figure~\ref{fig:4}). Other textual tokens obtained from each chart component are denoted as $\{w_{1,1}, ..., w_{1,m}\}$.

To capture semantic information about the chart, we incorporate a learnable one-hot embedding $z^{\rm typ}_k$ for chart component type.~The location of the k-th token within the chart is denoted by $x^{{\rm loc}}_k$, which is a 4-dimensional feature based on the relative bounding box coordinates as,
\begin{equation}
\nonumber
\tiny
x^{{\rm loc}}_k = (x_{k}^{{\rm min}}/W_{{\rm im}}, y_{k}^{{\rm min}}/H_{{\rm im}}, x_{k}^{{\rm max}}/W_{{\rm im}}, y_{k}^{{\rm max}}/H_{{\rm im}}),
\end{equation}
where $(x_{k}^{{\rm min}},y_{k}^{{\rm min}})$ and $(x_{k}^{{\rm max}},y_{k}^{{\rm max}})$ represent the coordinates of the top-left and bottom-right corners of the bounding box of the k-th token, while $W_{{\rm im}}$ and $H_{{\rm im}}$ represent the width and height of the chart, respectively. To take into account the appearance of each chart component, we concatenate the features of the center and corresponding keypoints to obtain the final appearance embedding $z^{{\rm app}}_k$.

\noindent \textbf{Data Variable Replacement Technique}.~We enhance the training process by incorporating the data variable substitution technique.~During training, the model generates the corresponding data variable instead of the actual value, avoiding errors and hallucinations that can result from treating data records as regular tokens.~For example, numerical values in the table cells are replaced with data variables such as ``product1," ``product2," etc.~This approach improves the accuracy and factual consistency of generated summaries, tables, and answers, particularly when multiple data records are involved.

To supervise the use of data variables, we introduce a new loss term that penalizes the model for generating tokens that do not match any data variable.~The loss term uses a function $D(x)$ that returns the data variable that matches the token $x$ if any and null otherwise.~The loss term is defined as follows,
\vspace{-2mm}
\begin{equation}
L_{\rm var} = - \frac{1}{T} \sum_t^{T} \sum_i^{N_t} \sum_{j \in V} I_{D(x_i) = v_j} \log P_{i,j}(t),
\vspace{-1mm}
\end{equation}
where $T$ is the length of the generated output, $N_t$ is the number of tokens in the $t$-th output, and $P_{i,j}(t)$ is the probability of generating the $i$-th token as the $j$-th data variable at the $t$-th time step. $I_{D(x_i) = v_j}$ is an indicator function that equals 1 if $D(x_i) = v_j$, and 0 otherwise.

The final optimization is to optimize the sum of the two losses, with $\alpha$ as a hyperparameter that balances the two losses,
\vspace{-1mm}
\begin{equation}
L = L_{\rm ans} + \alpha L_{\rm var},
\vspace{-1mm}
\end{equation}
where $\alpha$ is a hyperparameter that balances the two losses.~The loss  $L_{\rm ans}$ may be slightly adjusted depending on the specific chart understanding task.~In summary, our approach using data variable substitution can simultaneously support chart-to-text, chart-to-table, and chartQA tasks.~The introduced loss term also ensures the correct use of data variables during training.

\begin{table}[htbp]
\small
\caption{\small Datasets used for multiple chart understanding tasks, including Chart-to-Table, ChartQA, and Chart-to-Text.}
\vspace{-2mm}
\begin{center}
\renewcommand\arraystretch{1.1}
\begin{tabular}{llcc}
\toprule
\textbf{Tasks} & \textbf{Datasets} & \textbf{\#Charts} & \textbf{\#QAPs} \\
\midrule
Chart-to-Text  & C2T~\cite{obeid2020chart}               & 83K               & -               \\
Chart-to-Table & EC400K~\cite{luo2021chartocr}            & 387K              & -               \\
ChartQA        & FQA~\cite{kahou2017figureqa}               & 180K              & 2.4M            \\
ChartQA        & PlotQA~\cite{methani2020plotqa}               & 224K              & 28M             \\
ChartQA        & ChartQA~\cite{masry2022chartqa}           & 22K               & 33K             \\
ChartQA        & DVQA~\cite{kafle2018dvqa}              & 300K              & 3.5M \\
\bottomrule
\end{tabular}
\vspace{-5mm}
\end{center}
\label{tab:1}
\end{table}

\vspace{-4mm}
\section{Experiments}
\label{sec:experiments}
\vspace{-1mm}
\subsection{Evaluation Tasks and Datasets}
\vspace{-1mm}
We conducted experiments on multiple chart understanding tasks, including Chart-to-Table, ChartQA, and Chart-to-Text, as shown in Table~\ref{tab:1}.

\noindent \textbf{Chart-to-Text~Task}. For the Chart-to-Text task, we used the C2T~\cite{obeid2020chart} dataset, which contains two subsets: Pew and Statista. The Pew subset consists of chart images from Pew Research Center with automatically extracted summaries, while the Statista subset consists of chart images from Statista with human-written summaries. This dataset provides a diverse set of chart styles and textual summaries, enabling the development of effective chart-to-text models.

\noindent \textbf{Chart-to-Table~Task}. To evaluate the effectiveness of our approach in the Chart-to-Table task, we used the EC400K dataset~\cite{luo2021chartocr}. This dataset contains 386,966 real-world chart images from public Excel sheets and provides both bounding box locations and numerical readings of the charts. The EC400K dataset offers a wide variety of chart types and styles, surpassing previous datasets used in chart comprehension research. This enables us to validate the performance on diverse and challenging real-world chart data.

\noindent \textbf{ChartQA Task}. We evaluated our approach for ChartQA on four datasets: FQA~\cite{kahou2017figureqa}, DVQA~\cite{kafle2018dvqa}, PlotQA~\cite{methani2020plotqa}, and ChartQA~\cite{masry2022chartqa}. The FQA dataset includes charts with templates for binary answer questions and has two validation sets and two non-publicly available test sets. DVQA is a synthetic dataset that provides precise location and appearance of visual elements and metadata, including two test tasks: Test-Familiar and Test-Novel. PlotQA is a large and publicly available dataset for chart comprehension tasks, containing charts generated from real-world data. It provides two benchmarks, PlotQA-V1 and PlotQA-V2, with the latter including the former as a subset. ChartQA is a recent open-domain chart Q\&A dataset, evaluated on two subsets: augmented and human, where the augmented set is machine-generated and more extractive, while the human set is human-written and requires more complex reasoning.

\noindent \textbf{Evaluation Metrics}. We adopted task-specific metrics in our study. The Chart-to-Table task employed the metrics used in ChartOCR\cite{luo2021chartocr} to ensure fairness. For ChartQA, we used standard accuracy with a relaxed correctness criterion~\cite{masry2022chartqa,methani2020plotqa} that permits a maximum of\% numerical error. Additionally, BLEU4 was used to evaluate the Chart-to-Text task.

\begin{table*}[htbp]
\small
\caption{\small Performance comparison of different methods on the ChartQA task. The table shows the results of our proposed models, TaPas (Ours) and T5 (Ours), on four datasets: FQA, DVQA, PlotQA, and ChartQA. Our T5-based model achieved state-of-the-art performance on all datasets, outperforming previous methods by a large margin. The relaxed correctness criterion permits a maximum of 5\% numerical error. TF and TN stand for Test-Familiar and Test-Novel, respectively, for the DVQA dataset.}
\vspace{-2mm}
\begin{center}
\renewcommand\arraystretch{1.1}
\begin{tabular}{lcccccccccc}
\toprule
\textbf{}        & \multicolumn{4}{c}{\textbf{FQA}}                                  & \multicolumn{2}{c}{\textbf{DVQA}} & \multicolumn{2}{c}{\textbf{PQA}}    & \multicolumn{2}{c}{\textbf{ChartQA}} \\
\midrule
\textbf{Methods} & \textbf{Val1}  & \textbf{val2}  & \textbf{Test1} & \textbf{Test2} & \textbf{TF}       & \textbf{TN}   & \textbf{Test V1} & \textbf{Test V2} & \textbf{Val}      & \textbf{Test}    \\
\midrule
IMG+QUES~\cite{kafle2018dvqa}         & 59.41          & 57.14          & -              & 56.04          & 21.06             & 21.00         & -                & -                & -                 & -                \\
PReFIL~\cite{kafle2020answering}           & {\ul 94.84}    & {\ul 93.26}    & \textbf{94.88} & 93.16          & 69.66             & 69.53         & 57.91            & 10.37            & 4.53              & 4.80             \\
CRCT~\cite{levy2022classification}             & 94.61          & 85.04          & 94.23          & 84.77          & -                 & -             & {\ul 76.94}      & 34.44            & -                 & -                \\
PlotQA~\cite{methani2020plotqa}           & -              & -              & -              & -              & 57.99             & 59.54         & 53.96            & 22.52            & 36.15             & 38.00            \\
TaPas~\cite{masry2022chartqa}              & 90.32          & 90.43          & 89.52          & 89.57          & 48.82             & 48.68         & 15.09            & 12.90            & 39.68             & 41.28            \\
V-TaPas~\cite{masry2022chartqa}            & 91.46          & 91.45          & 90.68          & 90.64          & 94.43             & {\ul 94.54}   & 65.30            & 42.50            & 42.60             & 45.52            \\
T5~\cite{masry2022chartqa}                 & 87.97          & 87.83          & 87.56          & 87.57          & 89.01             & 89.01         & 72.62            & 56.22            & 40.15             & 41.04            \\
VL-T5~\cite{masry2022chartqa}            & 88.60          & 88.49          & 88.20          & 88.18          & 93.75             & 93.75         & 75.90            & 56.02            & 38.43             & 41.56            \\
TaPas (Ours)     & 91.12          & 91.40          & 91.15          & 91.25          & 92.20             & 94.30         & 74.20            & {\ul 56.20}      & {\ul 48.30}       & {\ul 50.20}      \\
T5 (Ours)        & \textbf{95.50} & \textbf{95.80} & {\ul 94.40}    & \textbf{93.40} & \textbf{95.40}    & \textbf{96.50}         & \textbf{78.10}   & \textbf{59.30}   & \textbf{49.50}    & \textbf{52.60} \\
\bottomrule
\end{tabular}
\end{center}
\label{tab:2}
\vspace{-6mm}
\end{table*}

\vspace{-2mm}
\subsection{Training Details for Different Tasks}
\label{sec:training-details}
\vspace{-1mm}
In this section, we provide an overview of the training details for our method on different tasks. The chart component detection model was trained on 8$\times$A6000 GPUs, while all other experiments based on pre-trained LLMs were fine-tuned using 64 GCP-TPUv3.

\noindent \textbf{Chart-to-Text Task}.~We fine-tuned the model for 10,000 steps with a learning rate of 2e-5, a batch size of 16, and a maximum sequence length of 512.

\noindent \textbf{Chart-to-Table Task}.~We trained the chart component detection module on the EC400K dataset\cite{luo2021chartocr}, which is currently the largest dataset for Chart-to-Table tasks. Ground truth labels for the position of the center and key points were derived from the bounding boxes labeled in the original dataset. During training, we used the Adam optimizer\cite{kingma2014adam} with a learning rate of 2.5e-4 and reduced the learning rate to 2.5e-5 for the last 5,000 batches, with a batch size of 32. Soft-NMS\cite{bodla2017soft} was used to merge key points from the heatmap. The hyper-parameters were set using a validation set, and early-stopping was used for end-to-end training. Although we did not fine-tune the chart component detection module on other tasks, the final model trained on other chart understanding tasks can still be used for extracting tabular data in Chart-to-Table tasks. The remaining settings were the same as in the previous work, ChartOCR~\cite{luo2021chartocr}.

\noindent \textbf{ChartQA Task}.~We performed fine-tuning for the ChartQA task on different datasets. For FigureQA, we used binary cross-entropy loss and the Adam optimizer with a base learning rate of 5e-4. The learning rate decayed by a factor of 0.7 from the 15th to 25th epoch. For DVQA, we used multinomial cross-entropy loss and the Adam optimizer with a base learning rate of 7e-4. The learning rate decayed by a factor of 0.6 from the 15th to 25th epoch. For PlotQA, we used binary cross-entropy loss and the Adam optimizer with a base learning rate of 5e-4. Negative examples were generated by randomly assigning wrong answers to questions, and the model was trained for 20 epochs with a linear learning rate scheduler.

\subsection{State-of-the-Art Comparisons}
\vspace{-1mm}
\noindent \textbf{Results of Chart-to-Text Task}. Table~\ref{tab:3} presents the results of the Chart-to-Text task, where we compare the performance of our T5-based model with previous state-of-the-art methods, PaLI-17B (res. 224) and Pix2Struct, on two datasets, Pew and Statista. Our T5-based model achieved the best performance on both datasets with significant improvements over the previous methods. By comparing our model with the original T5 model, we attribute our superior performance to the ability to generate diverse and informative summaries. It is worth noting that our T5-based model outperformed PaLI-17B (res. 588) on both datasets, despite having a smaller model size. This suggests that our approach can achieve good performance even with a smaller model size.

\begin{table}[htbp]
\small
\caption{\small~Results of Chart-to-Text task on the Pew and Statista datasets. Our T5-based model achieved state-of-the-art performance on both datasets, outperforming previous state-of-the-art methods.}
\vspace{-2mm}
\begin{center}
\renewcommand\arraystretch{1.1}
\begin{tabular}{lcc}
\toprule
\textbf{Methods}    & \textbf{Pew} & \textbf{Statista} \\
\midrule
T5~\cite{liu2022matcha}                  & 10.5         & 35.3              \\
PaLI-17B (res. 224)~\cite{liu2022matcha}  & 10.0         & 40.2              \\
PaLI-17B (res. 588)~\cite{liu2022matcha}  & 11.2         & 41.4              \\
Pix2Struct~\cite{lee2022pix2struct}          & 10.3         & 38.0              \\
MATCHA~\cite{liu2022matcha}               & 12.2         & 39.4              \\
T5 (Ours)           & \textbf{14.2}& \textbf{44.2}   \\
\bottomrule
\end{tabular}
\vspace{-6mm}
\end{center}
\label{tab:3}
\end{table}

\begin{table}[htbp]
\small
\caption{\small Comparison of ChartReader with previous state-of-the-art methods on EC400K for Chart-to-Table task. The reported GPU hrs refer to the total amount of GPU processing time used for evaluating on 8 $\times$A6000 GPUs, which is roughly equivalent to 8 times the running time, including time for IO and data preprocessing. Bold denotes the best performance.}
\vspace{-2mm}
\begin{center}
\renewcommand\arraystretch{1.1}
\begin{tabular}{lcccc}
\toprule
\textbf{Methods}   & \textbf{Bar}  & \textbf{Pie}  & \textbf{Line} & \textbf{GPU hrs} \\
\midrule
Revision~\cite{savva2011revision}            & 0.58          & 0.84          & -             & -                \\
Faster-RCNN~\cite{liu2019data, choi2019visualizing} & 0.80          & -             & -             & -                \\
Rotation-RNN~\cite{liu2019data} & -             & 0.80          & -             & -                \\
ChartOCR~\cite{luo2021chartocr}           & 0.92          & 0.92          & 0.96          & 57h             \\
Ours               & \textbf{0.95} & \textbf{0.95} & \textbf{0.97} & 22h  \\
\bottomrule
\end{tabular}
\vspace{-8mm}
\end{center}
\label{tab:4}
\end{table}

\noindent \textbf{Results of Chart-to-Table Task}. In Table~\ref{tab:4}, we report the results of our proposed method on the EC400K dataset~\cite{luo2021chartocr} for the Chart-to-Table task. Our approach achieved state-of-the-art performance, surpassing previous methods such as RotationRNN and Faster-RCNN, which rely solely on image classification and object detection. The comparison highlights the superiority of our key point detection approach over bounding box detection. Our approach also outperformed earlier attempts such as Revision and ChartOCR, which heavily rely on hand-crafted heuristic rules. Notably, ChartOCR represents the current state-of-the-art in the Chart-to-Table task. Our superior performance can be attributed to two factors. Firstly, we eliminated the need for heuristic rules and learned to recognize chart components by grouping center points and key points. Secondly, the recognized chart components are further utilized in subsequent chart-to-table and chartQA tasks, allowing our model to better understand the structure and semantic information of charts, leading to improved numerical evaluation performance. Moreover, our method achieved this superior performance with significantly less GPU hours compared to ChartOCR, as shown in Table~\ref{tab:2}. Specifically, our method only required 22 GPU hours, while ChartOCR used 57 GPU hours. This indicates that our method not only outperforms previous approaches but also does so with more efficient use of computing resources.

\begin{table}[htbp]
\small
\caption{\small Ablation study on chart component detection, comparing the performance of the full model with two variants that do not have either CT point detection or group detection. The full model outperforms the ablated models on all three chart types, highlighting the importance of both components for accurate chart component detection.}
\vspace{-2mm}
\begin{center}
\renewcommand\arraystretch{1.1}
\begin{tabular}{lccc}
\toprule
\textbf{Methods} & \textbf{Bar} & \textbf{Pie} & 
\textbf{Line} \\
\midrule
w/o CT Point     & 0.83         & 0.73         & 0.63          \\
w/o Group        & 0.77         & 0.81         & 0.52          \\
Ours             & 0.95         & 0.95         & 0.97     \\
\bottomrule
\end{tabular}
\vspace{-6mm}
\end{center}
\label{tab:5}
\end{table}

\begin{table}[htbp]
\small
\caption{\small Ablation study on input encoding for the PlotQA-V1 and PlotQA-V2 datasets. The table shows that the full model outperforms all ablated versions, and the location and appearance embeddings contribute the most to the overall performance.}
\vspace{-2mm}
\begin{center}
\renewcommand\arraystretch{1.1}
\begin{tabular}{lcc}
\toprule
\textbf{Ablation Token} & \textbf{PlotQA-V1} & \textbf{PlotQA-V2} \\
\midrule
w/o type                & 60.20              & 48.20              \\
w/o location            & 64.20              & 51.20              \\
w/o appearance          & 59.10              & 45.20              \\
w/o CCD                 & 52.30              & 40.20              \\
Ours (TaPas)            & 74.20              & 56.20  \\
\bottomrule
\end{tabular}
\vspace{-8mm}
\end{center}
\label{tab:6}
\end{table}

\noindent \textbf{Results of ChartQA Task}.~Table \ref{tab:2} shows the results of our proposed models, TaPas (Ours) and T5 (Ours), on all datasets in the ChartQA task. Our approach utilizes the structural and semantic information of charts to answer questions, making it highly suitable for complex chart understanding tasks such as ChartQA. Specifically, our T5-based model outperformed the previous state-of-the-art method, PReFIL, on the FQA dataset by a small margin, and achieved state-of-the-art performance on the other three datasets. Our model benefits from joint training on chart comprehension and extraction tasks, enabling it to better understand the semantic and structural information of charts. Our T5-based model achieved state-of-the-art performance on the DVQA and PlotQA datasets, outperforming previous methods, including the current state-of-the-art method, V-TaPas, on both test sets. Additionally, our T5-based model achieved state-of-the-art performance on the ChartQA dataset, outperforming previous methods by a large margin, indicating good generalization to unseen data.

\begin{table}[htbp]
\small
\caption{\small Impact of hyperparameter $\alpha$ in Data Variable Replacement Technique.}
\vspace{-2mm}
\begin{center}
\renewcommand\arraystretch{1.1}
\begin{tabular}{lccccc}
\toprule
\textbf{}             & \textbf{FQA} & \textbf{DVQA} & \textbf{PlotQA} & \textbf{CharQA} & \textbf{C2T} \\
\midrule
 $\alpha$ & 0.3          & 0.2           & 0.6             & 0.7             & 0.5          \\
Rate                  & 5\%          & 5\%           & 20\%            & 40\%            & 30\%       \\
\bottomrule
\end{tabular}
\vspace{-6mm}
\end{center}
\label{tab:7}
\end{table}

\begin{table}[htbp]
\small
\caption{\small Results of ablation study on the effect of pre-training dataset on the ChartQA task.}
\vspace{-4mm}
\begin{center}
\renewcommand\arraystretch{1.1}
\begin{tabular}{cllccc}
\toprule
\textbf{Model} & \textbf{PT}      & \textbf{EF} & \textbf{MT} & \textbf{Val} & \textbf{Test} \\
\midrule
T5            & C4+(PQA)         & CQA     & \ding{55}        & 38.4         & 39.2          \\
T5            & C4+(PQA)         & CQA     & \ding{52}        & 42.3         & 41.5          \\
T5            & C4+(PQA+CQA)     & CQA     & \ding{55}        & 39.2         & 39.5          \\
T5            & C4+(PQA+CQA)     & CQA     & \ding{52}        & 45.5         & 43.2          \\
T5            & C4+(PQA+CQA+C2T) & CQA     & \ding{55}        & 41.2         & 42.2          \\
T5            & C4+(PQA+CQA+C2T) & CQA     & \ding{52}        & 49.5         & 52.6  \\
\bottomrule
\end{tabular}
\vspace{-8mm}
\end{center}
\label{tab:8}
\end{table}

\vspace{-1mm}
\section{Ablation Study}
\vspace{-1mm}
\noindent \textbf{Chart Component Detection}.~We compare our full model with two variants that do not have either CT point detection or group detection. As shown in the Table~\ref{tab:5}, the full model outperforms the ablated models on all three chart types, and the CT point module improves performance on line charts, while the group module is more effective for bar charts. These results highlight the importance of both components for accurate chart component detection.

\noindent \textbf{Input Encoding}.~We perform an ablation study on the input encoding method, comparing the performance of our full model and different ablated versions on PlotQA-V1 and PlotQA-V2 datasets.~Table~\ref{tab:6} shows that the full model significantly outperforms all ablated versions, and the location and appearance embeddings contribute the most to the overall performance. This indicates that spatial and visual information is critical for chart comprehension tasks.

\noindent \textbf{Hyperparameter $\alpha$ in Data Variable Replacement}.~In this section, we conduct an ablation study to investigate the impact of the hyperparameter $\alpha$ in the Data Variable Replacement Technique on five datasets.~As shown in Table~\ref{tab:7},~the optimal value of $\alpha$ varies across datasets, ranging from 0.2 to 0.7.~We speculate that more complex tasks require a higher weight on replaced data variables. Our results demonstrate the effectiveness of the data variable replacement technique.

\noindent \textbf{Multi-Task~Pre-Training}.~We conduct ablation experiments to evaluate the effectiveness of our proposed chart component detection and multi-task training approach on the ChartQA task.~As shown in the Table~\ref{tab:8}, our model trained on the C4+ pre-training dataset with all three chart-related tasks achieves the best performance on both validation and test sets. These results demonstrate the importance of incorporating multi-task training.

\vspace{-1mm}
\vspace{-1mm}
\section{Conclusion}
\vspace{-2mm}
In this paper, we have proposed ChartReader, a unified framework for chart comprehension that integrates chart derendering and comprehension tasks seamlessly. Our approach leverages a transformer-based chart component detection module and an extended pre-trained vision-language model for chart comprehension tasks, eliminating the need for manual rule-making and enhancing accuracy.~Through extensive experiments, we have demonstrated that ChartReader outperforms existing methods in~Chart-to-Table,~ChartQA, and~Chart-to-Text tasks.~Our framework has the potential to significantly reduce manual effort in chart analysis and facilitate a step towards a universal chart understanding model.

\clearpage
\setcounter{section}{0}
\setcounter{figure}{0}
\setcounter{table}{0}
\renewcommand\thesection{\Alph{section}}
\renewcommand\thesubsection{\thesection.\arabic{subsection}}
\renewcommand\thefigure{A\arabic{figure}}

\section*{Supplementary materials}

\begin{figure*}
\small
    \centering
    \includegraphics[width=0.95\linewidth]{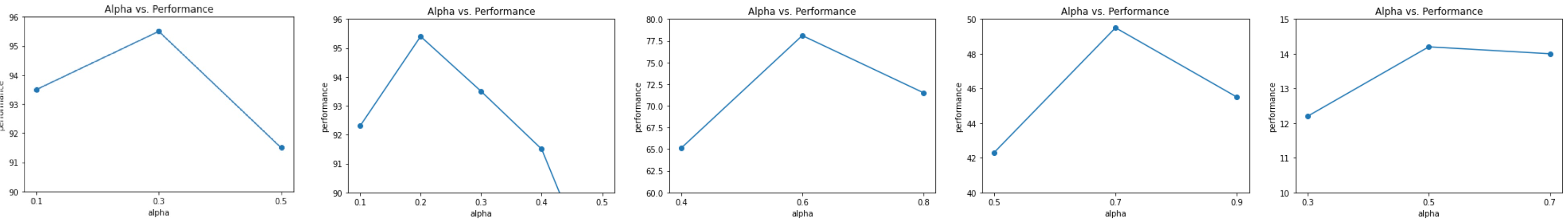}
    \caption{\small shows the impact of the hyperparameter $\alpha$ on model performance in chart understanding. The experimental results indicate that the weight of the $\alpha$ parameter increases as the number of open-ended questions grows, as the model needs to use variable replacement techniques more frequently to avoid errors caused by random guessing of unknown markers. Appropriate selection of the $\alpha$ parameter is crucial for achieving high performance and training efficiency of pretrained models.}
    \label{fig:alpha}
\end{figure*}

\noindent \textbf{Discussion on the Hyperparameter $\alpha$}. The hyperparameter $\alpha$ is crucial for achieving high performance and training efficiency of pretrained models. It is used in the data variable replacement technique to control how the model handles unknown markers in the input data. We conducted additional experiments to investigate the impact of the alpha parameter, as shown in Figure~\ref{fig:alpha}. The experimental results indicate that the weight of the $\alpha$ parameter increases as the number of open-ended questions grows, as the model needs to use variable replacement techniques more frequently to avoid errors caused by random guessing of unknown markers. However, selecting an inappropriate value for the $\alpha$ parameter can make the model difficult to converge. Therefore, selecting the value of the alpha parameter requires careful consideration to maximize the performance and training efficiency of the model.

\noindent \textbf{Discussion on Training Steps}. In Table~\ref{tab:pretraining}, we have analyzed the optimal sequence for training the QA or Text task and found that training the QA task last led to improved performance, especially when the QA dataset was included. Similarly, we observed that training the Chart to Text task last also resulted in better performance. We also explored the order in which the various datasets within the QA task should be trained and found that training from easy to hard produced better results. However, we discovered that multi-task training, where all datasets were combined, resulted in even better performance, possibly due to the inherent complementarity between the QA and Text tasks. Additionally, we considered the scale of different datasets and provided a fusion weight, which is shown in the "Rate" of Table~\ref{tab:alpha}.

\noindent \textbf{Comparison of Different Model Variants}. Our proposed method enhances the performance of different models, as seen in our comparison of the T5 model with other variants such as T5S. Incorporating the CCD module and MT significantly improves the model's generalization performance and stability compared to the original T5 baseline, as presented in Table~\ref{tab:add-exp}. However, including the CCD module slightly reduces training efficiency due to additional training time and I/O operations. Nonetheless, optimizing these operations would not significantly increase the training time, and the increase in TPU hours remains within an acceptable range. We remain committed to continuing research in this area and further optimizing the proposed method to enhance its effectiveness.

\noindent \textbf{Discussion on Limitations}. While our proposed method has shown promising results in handling multiple tasks and supporting cross-dataset and cross-task learning, the performance of the model may be affected by the size and quality of the pretraining dataset and the specific fine-tuning task. Selecting appropriate hyperparameters can also be challenging and may require extensive experimentation. These limitations suggest that further research is needed to optimize our proposed method and improve its performance. Nonetheless, our work provides valuable insights into the development of a universal chart understanding framework.

\begin{table}[htbp]
\small
\caption{\small Impact of hyperparameter $\alpha$ in Data Variable Replacement Technique. The QA and Text tasks are divided into sub-datasets of varying difficulty levels, and the "Rate" column shows the fusion weight used for multi-task training.}
\vspace{-2mm}
\begin{center}
\renewcommand\arraystretch{1.1}
\begin{tabular}{lccccc}
\toprule
\textbf{}             & \textbf{FQA} & \textbf{DVQA} & \textbf{PlotQA} & \textbf{CharQA} & \textbf{C2T} \\
\midrule
 $\alpha$ & 0.3          & 0.2           & 0.6             & 0.7             & 0.5          \\
Rate                  & 5\%          & 5\%           & 20\%            & 40\%            & 30\%       \\
\bottomrule
\end{tabular}
\vspace{-6mm}
\end{center}
\label{tab:alpha}
\end{table}

\begin{table}[htbp]
\small
\caption{\small Pretraining Dataset and Training Steps. This table shows the details of the combined pretraining dataset and the training steps involved in the different tasks, including QA, Text, and Chart to Text tasks. The table also includes the fusion weights used for combining the different datasets within the QA task.}
\vspace{-2mm}
\begin{center}
\renewcommand\arraystretch{1.1}
\begin{tabular}{lcccc}
\toprule
\textbf{}                     & \multicolumn{2}{l}{\textbf{ChartQA}} & \multicolumn{2}{l}{\textbf{Chart-to-Text}} \\
\midrule
\textbf{Experimental Setting} & \textbf{Val}      & \textbf{Test}    & \textbf{Pew}       & \textbf{Statista}     \\
\midrule
QA-\textgreater{}Text         & 46.2              & 47.2             & 13.4               & 43.2                  \\
Text-\textgreater{}QA         & 48.2              & 51.6             & 12.7               & 41.2                  \\
QA by difficulty              & 48.3              & 51.2             & -                  & -                     \\
QA with random order          & 47.5              & 50.4             & -                  & -                     \\
Mixed training                & \textbf{49.5}     & \textbf{52.6}    & \textbf{14.2}      & \textbf{44.2}       \\
\bottomrule
\end{tabular}
\vspace{-6mm}
\end{center}
\label{tab:pretraining}
\end{table}

\begin{table*}[htbp]
\footnotesize
\caption{\small Comparison of performance among different model variants. The results show that incorporating the CCD module and MT significantly improves the generalization performance and stability of the model compared to the original T5 baseline. The proposed method enhances the performance of different models, as seen in the comparison of the T5 model with other variants such as T5S. The table also shows the increase in training time and TPU hours.}
\vspace{-2mm}
\begin{center}
\renewcommand\arraystretch{1.1}
\begin{tabular}{lllccccccccc}
\toprule
\textbf{}      & \textbf{}        & \textbf{}   & \textbf{}     & \textbf{}    & \multicolumn{2}{c}{\textbf{PQA (PlotQA )}} & \multicolumn{2}{c}{\textbf{CQA (ChartQA)}} & \multicolumn{2}{c}{\textbf{C2T (Table-to-Text)}} & \textbf{}        \\
\textbf{Model} & \textbf{PT}      & \textbf{EF} & \textbf{CCD.} & \textbf{MT.} & \textbf{Test V1}     & \textbf{Test V2}    & \textbf{Val}         & \textbf{Test}       & \textbf{Pew}          & \textbf{Statista}        & \textbf{TPU hrs} \\
\midrule
T5             & C4               & SQuAD       & \ding{55}        & \ding{55}       & 31.3                 & 21.2                & 16.9                 & 18.0                & 6.9                   & 22.8                     & -                \\
\midrule
T5             & C4+(CQA)         & PQA         & \ding{55}        & \ding{55}       & 65.7                 & 42.9                & 23.4                 & 22.3                & -                     & -                        & 17               \\
T5             & C4+(CQA)         & PQA         & \ding{52}        & \ding{55}       & 68.4                 & 51.7                & 26.4                 & 24.2                & -                     & -                        & 31               \\
T5             & C4+(CQA)         & PQA         & \ding{55}        & \ding{52}       & 69.5                 & 52.5                & 28.3                 & 27.2                & -                     & -                        & 21               \\
T5             & C4+(CQA)         & PQA         & \ding{52}       & \ding{52}       & 73.5                 & 56.5                & 32.6                 & 31.6                & -                     & -                        & 37               \\
\midrule
T5             & C4+(PQA)         & CQA         & \ding{55}        & \ding{55}       & 45.6                 & 24.8                & 38.4                 & 39.2                & -                     & -                        & 13               \\
T5             & C4+(PQA)         & CQA         & \ding{52}        & \ding{55}       & 53.9                 & 36.6                & 38.7                 & 39.6                & -                     & -                        & 28               \\
T5             & C4+(PQA)         & CQA         & \ding{55}        & \ding{52}       & 56.8                 & 40.8                & 37.6                 & 36.6                & -                     & -                        & 19               \\
T5             & C4+(PQA)         & CQA         & \ding{52}       & \ding{52}       & 63.8                 & 47.6                & 42.3                 & 41.5                & -                     & -                        & 35               \\
\midrule
T5             & C4+(CQA+C2T)     & PQA         & \ding{55}        & \ding{55}       & 31.8                 & 21.8                & 27.4                 & 26.2                & -                     & -                        & 24               \\
T5             & C4+(CQA+C2T)     & PQA         & \ding{52}        & \ding{55}       & 68.6                 & 45.9                & 30.3                 & 36.4                & -                     & -                        & 36               \\
T5             & C4+(CQA+C2T)     & PQA         & \ding{55}        & \ding{52}       & 70.5                 & 53.5                & 29.4                 & 38.4                & -                     & -                        & 27               \\
T5             & C4+(CQA+C2T)     & PQA         & \ding{52}       & \ding{52}       & 74.6                 & 55.2                & 34.2                 & 31.5                & -                     & -                        & 41               \\
\midrule
T5             & C4+(PQA+C2T)     & CQA         & \ding{55}        & \ding{55}       & 48.6                 & 18.8                & 38.3                 & 36.4                & -                     & -                        & 22               \\
T5             & C4+(PQA+C2T)     & CQA         & \ding{52}        & \ding{55}       & 52.8                 & 35.6                & 27.3                 & 40.3                & -                     & -                        & 35               \\
T5             & C4+(PQA+C2T)     & CQA         & \ding{55}        & \ding{52}       & 54.6                 & 36.9                & 38.5                 & 39.2                & -                     & -                        & 25               \\
T5             & C4+(PQA+C2T)     & CQA         & \ding{52}       & \ding{52}       & 65.7                 & 49.7                & 45.5                 & 44.3                & -                     & -                        & 38               \\
\midrule
T5             & C4+(ALL)& PQA         & \ding{55}        & \ding{55}       & 61.2                 & 43.4                & 29.3                 & 31.5                & -                     & -                        & 28               \\
T5             & C4+(ALL)& PQA         & \ding{52}        & \ding{55}       & 72.2                 & 52.5                & 36.2                 & 34.4                & -                     & -                        & 41               \\
T5             & C4+(ALL)& PQA         & \ding{55}        & \ding{52}       & 73.3                 & 54.2                & 35.4                 & 33.3                & -                     & -                        & 32               \\
T5             & C4+(ALL)& PQA         & \ding{52}       & \ding{52}       & \textbf{78.1}        & \textbf{59.3}       & 39.4                 & 37.4                & -                     & -                        & 47               \\
\midrule
T5             & C4+(ALL)& CQA         & \ding{55}        & \ding{55}       & 43.8                 & 15.7                & 41.2                 & 42.2                & -                     & -                        & 26               \\
T5             & C4+(ALL)& CQA         & \ding{52}        & \ding{55}       & 48.1                 & 31.6                & 44.4                 & 46.5                & -                     & -                        & 38               \\
T5             & C4+(ALL)& CQA         & \ding{55}        & \ding{52}       & 52.7                 & 34.8                & 45.4                 & 45.4                & -                     & -                        & 27               \\
T5             & C4+(ALL)& CQA         & \ding{52}       & \ding{52}       & 72.7                 & 53.5                & \textbf{49.5}        & \textbf{52.6}       & -                     & -                        & 44               \\
\midrule
T5             & C4+(ALL)& C2T         & \ding{55}        & \ding{55}       & -                    & -                   & -                    & -                   & 9.6                   & 34.7                     & 27               \\
T5             & C4+(ALL)& C2T         & \ding{52}        & \ding{55}       & -                    & -                   & -                    & -                   & 12.5                  & 37.5                     & 42               \\
T5             & C4+(ALL)& C2T         & \ding{55}        & \ding{52}       & -                    & -                   & -                    & -                   & 13.2                  & 38.8                     & 34               \\
T5             & C4+(ALL)& C2T         & \ding{52}       & \ding{52}       & -                    & -                   & -                    & -                   & \textbf{14.2}         & \textbf{44.2}            & 44               \\
\midrule
Tapas          & C4+(ALL)& PQA         & \ding{52}       & \ding{52}       & 74.4                 & 56.5                & 37.5                 & 36.5                & -                     & -                        & 54               \\
Tapas          & C4+(ALL)& CQA         & \ding{52}       & \ding{52}       & 48.7                 & 50.6                & 48.8                 & 50.7                & -                     & -                        & 47               \\
Tapas          & C4+(ALL)& C2T         & \ding{52}       & \ding{52}       & -                    & -                   & -                    & -                   & 12.8                  & 38.8                     & 43               \\
\midrule
T5S            & C4+(ALL)& PQA         & \ding{52}       & \ding{52}       & 72.3                 & 55.3                & 34.5                 & 32.4                & -                     & -                        & 34               \\
T5S            & C4+(ALL)& CQA         & \ding{52}       & \ding{52}       & 64.3                 & 47.3                & 44.4                 & 43.4                & -                     & -                        & 32               \\
T5S            & C4+(ALL)& C2T         & \ding{52}       & \ding{52}       & -                    & -                   & -                    & -                   & 12.4                  & 36.7                     & 31    \\
\bottomrule
\end{tabular}
\end{center}
\label{tab:add-exp}
\vspace{-6mm}
\end{table*}

\section*{ACKNOWLEDGMENTS}
This work received support from the Air Force Research Laboratory under agreement number FA8750-19-2-0200; the Defense Advanced Research Projects Agency (DARPA) grants funded through the GAILA program (award HR00111990063); and the Defense Advanced Research Projects Agency (DARPA) grants funded through the AIDA program (FA8750-18-20018).

The U.S. Government is authorized to reproduce and distribute reprints for Governmental purposes notwithstanding any copyright notation thereon. The views and conclusions contained herein are those of the authors and should not be interpreted as necessarily representing the official policies or endorsements, either expressed or implied, of the Air Force Research Laboratory or the U.S. Government.

{\small
\bibliographystyle{ieee_fullname}
\bibliography{egbib}
}

\end{document}